\documentclass[10pt,twocolumn,letterpaper]{article}

\usepackage{iccv}
\usepackage{times}
\usepackage{epsfig}
\usepackage{graphicx}
\usepackage{amsmath}
\usepackage{amssymb}
\usepackage{caption}
\usepackage{lipsum}
\usepackage{bbm}
\usepackage{booktabs}
\usepackage{cuted}
\usepackage{array} 
   \usepackage{url}
   \usepackage[accsupp]{axessibility}

\usepackage[pagebackref=true,breaklinks=true,letterpaper=true,colorlinks,bookmarks=false]{hyperref}

 \iccvfinalcopy 

\def\iccvPaperID{} 

\ificcvfinal\fi

\begin{document}

\title{Automatic Animation of Hair Blowing in Still Portrait Photos}

\author{Wenpeng Xiao$^{1}$, Wentao Liu$^{1}$, Yitong Wang$^{1}$, Bernard Ghanem$^{2}$, Bing Li$^{2,*}$ \vspace{3pt} \\
 $^1$ ByteDance Intelligent Creation Lab \\ $^2$ King Abdullah University of Science and Technology (KAUST)\\
{\small \{xiaowenpeng.com, liuwentao.canon\}@bytedance.com, wangyitong@pku.edu.cn, \{Bernard.Ghanem, bing.li\}@kaust.edu.sa}
}

 \twocolumn[{%
\renewcommand\twocolumn[1][]{#1}%
\maketitle
\begin{center}
    \centering
    \captionsetup{type=figure}
   \includegraphics[width=0.85\textwidth]{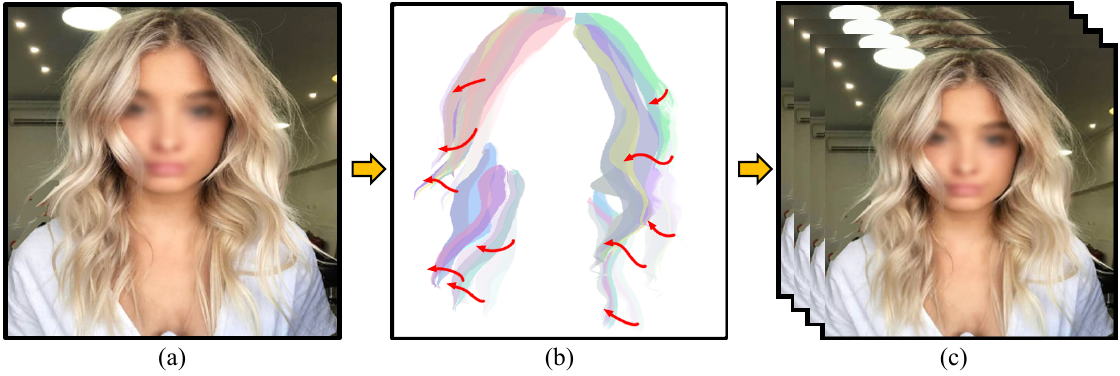}
   \vspace{5pt}
       \captionof{figure}{
    Given a still portrait photo (a), our method automatically detects hair wisps and animates wisps (b), while converting the photo into a cinemagraph (c). 
    }
    \label{fig:teaser}
\end{center}%
}]

\maketitle
\ificcvfinal\thispagestyle{empty}\fi

\begin{abstract}
   We propose a novel approach to animate human hair in a still portrait photo. Existing work has largely studied the animation of fluid elements such as water and fire. However,  hair animation for a real image remains underexplored, which is a challenging problem, due to the high complexity of hair structure and dynamics. Considering the complexity of hair structure, we innovatively treat hair wisp extraction as an instance segmentation problem, where a hair wisp is referred to as an instance. With advanced instance segmentation networks, our method extracts meaningful and natural hair wisps.  Furthermore, we propose a wisp-aware animation module that animates hair wisps with pleasing  motions without noticeable artifacts. 
   The extensive experiments show the superiority of our method. Our method provides the most pleasing and compelling viewing experience in the qualitative experiments, and outperforms state-of-the-art still-image animation methods by a large margin in the quantitative evaluation.  
   Project url: \url{https://nevergiveu.github.io/AutomaticHairBlowing/}
\end{abstract}

\vspace{-7pt}
  

\let\thefootnote\relax\footnotetext{$^*$ Corresponding Author.}
\section{Introduction}

"His silvery hair was blowing in the wind,”
George R.R. Martin\footnote{Celebrated novelist  of  "Wild Cards",  "A Game of Thrones", etc. \cite{RRGeorge}} described. 

Hair is one of the most impressive parts of the human body \cite{reed1990influence,wogalter1991effects}, while its dynamics make a deeper  impression  and make the scene vivid.  The studies show that dynamics is more compelling and captivating than a still image. 
Massive  portrait photos are shared every day on social media platforms such as TikTok and Instagram. People want their photos to be attractive and artistic. This motivates us to explore animating human hair  in a still image, so as to provide a vivid, pleasing and beautiful viewing experience. 
Recent methods \cite{endoSA2019,mahapatra2021controllable,Holynski_2021_CVPR} have been proposed to augment a still image with dynamics, which animates fluid elements such as water, smoke and fire in the image. However, these methods haven't taken   human hair into account for real photos.

To provide an artistic effect,  we focus on animating human hair in a portrait photo, while translating the photo into a \textbf{\textit{cinemagraph}} \cite{tompkin2011towards,cinemagraphweb}. Cinemagraph is an innovative short-video format preferred by  professional photographers, advertisers, and artists, and it is used in digital advertisements, social media,  landing pages, etc.
The engaging  nature of the cinemagraph is that  it integrates the merits of still photos and videos \cite{oh2017personalized,mahapatra2021controllable,bai2013automatic}. That is, some regions in a cinemagraph contain small  motions in a short loop,  while the rest remain static.  The contrast between static and moving elements helps to capture viewers' attention. Translating a portrait photo  into a cinemagraph with subtle hair motions would make the photo to be more compelling   yet  not distract viewers from static content.

Existing methods (\eg \cite{oh2017personalized,yan2017turning}) and commercial software (\cite{cinemagraphsoft1,cinemagraphsoft2,cinemagraphsoft3}) generate a high-fidelity cinemagraph from an input video by freezing  selective  video regions. These tools are not applicable to a still image.  On the other hand, still-image animation has attracted increasing attention \cite{endoSA2019,mahapatra2021controllable,Holynski_2021_CVPR}. 
Most approaches explore animating fluid elements such as clouds, water, and smoke.  However, hair is made of  fibrous materials, leading to its dynamics  being rather different from that of  fluid elements. Different from fluid element animation which has been largely investigated, human hair animation is much less been explored for a real portrait photo. 

Animating hair in  a still portrait photo is a  challenging problem.  
Research on hair modeling, which aims to reconstruct plausible hair for virtual humans, has revealed the high complexity of hair in terms of structure and  dynamics. 
For example, different from  human body or face that has smooth surfaces,   hair comprises around hundred-thousand  components \ie \textit{hair strands},  leading to intricate structures.  Furthermore,  such a  massive number of  fibrous components result in  non-uniform and  complicated   motions within the hair as well as temporal collisions  between hair and head.
Many hair modeling approaches address hair complexity by  resorting to specialized hair capture techniques (\eg dense camera array and high-speed cameras). Thus, static hair modeling approaches \cite{hu2014robust,luo2013structure,nam2019strand,rosu2022neuralstrands} construct high-quality 3D  models of static hair,  and dynamic hair modeling approaches  \cite{Wang_2022_CVPR,xu2014dynamic}  achieve   impressive results in reconstructing hair motions at strand level. 
However, these approaches suffer from expensive time costs or  rely on complex hardware setups to capture real-world hair.

In this paper, we propose a novel method that automatically  animates hair in a still  portrait photo without any user assistance or  sophisticated hardware. We observe  that  human visual system is much less sensitive to hair strands and their motions in a real portrait video,  compared to synthetic strands of a digitized human in a virtual environment.  Our insight is that we can animate hair wisps rather than strands which can create a perceptually pleasing viewing experience.  We hence propose a hair wisp animation module to  animate hair wisps, enabling an efficient solution.

The arising  challenge is  how to extract hair wisps.  
Although  relevant work such as hair modeling investigates hair segmentation,  these approaches focus on extracting the whole hair region, which is different from our aim. 
To  extract meaningful hair wisps, we innovatively  treat hair wisp extraction as  an instance segmentation problem, where a segment instance from a still image is referred to as a hair wisp. Thanks to such a problem definition, we can exploit instance segmentation networks to realize the extraction of  hair wisps. This largely simplifies the  hair wisp extraction problem, but also advanced networks can effectively  extract hair wisps. Furthermore, we construct a hair wisp dataset that consists of real portrait photos  to train the networks. We also propose a semi-annotation scheme to produce ground-truth annotations of hair wisps.

Our contributions  are summarized as follows:

\begin{itemize}
\item We propose a novel approach that automatically animates blowing hair from a still portrait image. Our method effectively handles high-resolution images, while generating   high-quality and  aesthetically-pleasing cinemagraphs without any user assistance.   
\item We show that  instance segmentation facilitates the animation of hair blowing, which is helpful in generating realistic  blowing motions.  
\item We propose a hair wisp animation module that generates pleasing motions for hair wisps without noticeable  artifacts. 
\end{itemize}

\section{Related Work}

\begin{figure*}[h]
\centering
\includegraphics[width=1\textwidth]{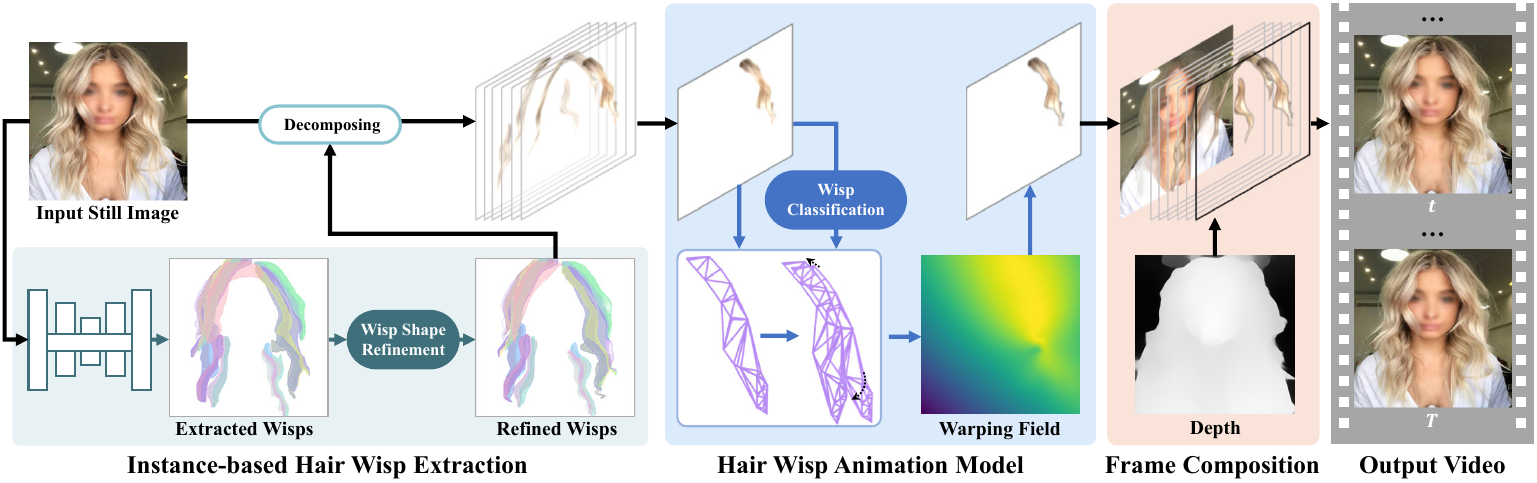}
\caption{The framework of our method. Given a still photo, our instance-based hair wisp extraction module extracts hair wisps from it. The proposed hair wisp animation module then represents the extracted hair wisps  by multi-layer mesh representation   and creates motions for each extracted wisp by driving the wisp's mesh.  Animated hair wisps and the background are composited  to render  frame $\Tilde{I}^t$ for generating a  video using the depth information of the photo.     }
\label{fig:framework}
\end{figure*}

\textbf{Hair Modeling.} Many efforts  have been devoted to hair modeling which is to generate/reconstruct human hairs  for virtual  humans. Hair modeling from scratch is laborious and time-consuming. Modeling methods \cite{choe2005statistical,kim2002interactive,yuksel2009hair} are proposed to synthesize  hair strands  to ease manual work. Detailed discussions can be found  in  the  survey\cite{ward2007survey}.   
To  narrow down  the gap between synthetic and real  hair,  hair capture methods are introduced to construct hair models from the real world. These methods  can be roughly categorized into static hair capture and dynamic one.  
Static hair capture methods  model static hair using various sensors. Methods \cite{paris2008hair,jakob2009capturing,beeler2012coupled,luo2013structure,nam2019strand,rosu2022neuralstrands,paris2008hair,kuang2022deepmvshair}  deploy  3D hair models from multi-view images or point clouds. 
However, most hair capture methods require  3D acquisition devices while   relying on complex hardware setups. 
Instead, a few work   has explored  reconstructing hair   from a single-view image \cite{chai2016autohair,chai2012single,hu2015single,chai2012single,zhou2018hairnet,chai2013dynamic,zhang2019hair,wu2022neuralhdhair} or sparse views  \cite{kuang2022deepmvshair}.

A few work \cite{zhang2012simulation,xu2014dynamic,yang2019dynamic,Wang_2022_CVPR,hu2017simulation,daviet2011hybrid} has explored dynamic hair capture. 
Most methods \cite{zhang2012simulation,xu2014dynamic, Wang_2022_CVPR,hu2017simulation} employ specialized hardware  techniques such as calibrated camera array and lighting control to  capture hair dynamics, where Wang \etal \cite{Wang_2022_CVPR} propose a hair-tracking algorithm  using
multi-view videos. Zhang \etal \cite{zhang2012simulation} and  Hu \etal \cite{hu2017simulation} exploit physical simulation for hair modeling.
 Winberg \etal \cite{winberg2022facial} employ 14 synchronized cameras to  track  3D facial hair and the underlying 3D skin.
 Yang \etal \cite{yang2019dynamic}  introduce the first deep-learning networks   for  dynamic hair modeling, where  synthetic hair data   are used to train the networks.
Different from  these hair modeling approaches designed for virtual humans, our work focuses on automatically animating hair for a real portrait photo.

\textbf{Single-image-to-video generation. }
Different video-based synthesis methods (\eg \cite{bai2013automatic,liao2013automated,oh2017personalized,tompkin2011towards,yeh2012tool})  generating video from the input video,  many  methods \cite{endoSA2019,mahapatra2021controllable,chuang2005animating,Holynski_2021_CVPR,geng2018warp,averbuch2017bringing,blattmann2021ipoke,castrejon2019improved,Dorkenwald_2021_CVPR,weng2019photo,hornung2007character,ni2023conditional} have been proposed to convert a still image into a video. As  generative models have shown impressive performance in  image and video  translation, video prediction methods \cite{blattmann2021ipoke,castrejon2019improved,Dorkenwald_2021_CVPR,hu2023dmvfn}  predict video frames from a single one using  Generative Adversarial Networks (GANs), variational autoencoders,  conditional invertible neural
network. These methods often suffer low-resolution issues, since understanding and generating videos are challenging \cite{xu2020g,zhao2021video}.  
 Differently, methods  \cite{chuang2005animating,hao2018controllable} animate a still image by estimating  motion fields.  
 With manually segmented layers, Chuang  et al. \cite{chuang2005animating} animate a layer with  a stochastic motion texture using harmonic oscillations \cite{sun2003video}.
Hao \etal \cite{hao2018controllable}  generate a video  from  an image while taking additional sparse motion trajectories as input,  where motion trajectories are provided by  users.  Halperin \etal combine conditional random field with a local feature descriptor to  calculate dense displacement fields from a user-provided motion direction.  Mahapatra and Kulkarni \cite{mahapatra2021controllable} generate a dense optical flow map from  user guidance   by exponential functions and GANs. 
  These methods require user assistance such as motion directions or a manual mask indicating  which regions need to be animated.
 Recent work  \cite{endoSA2019,Holynski_2021_CVPR} design motion estimation networks to predict motion field. Holynski \etal \cite{Holynski_2021_CVPR}  focus on  animating fluid elements, where the motions of a generated video are represented by a static motion field. 
Methods \cite{geng2018warp,averbuch2017bringing,li2021anigan,xu2022region} have devoted efforts to  portrait image animation, where some methods  \cite{xia2021tedigan,xu2022region} utilize attribute-level information \cite{liang2018visual,liang2018unifying}.
However, these approaches mainly focus on  editing facial images,  but also rely on additional  reference videos or 3D face models.
Different from these approaches, our method focuses on animating human  hair in a still image without user assistance.

\section{Methodology}
\textbf{Problem Definition.}
Given a still portrait  photo $I$, our aim is to automatically generate  a cinemagraph  $V=\{\Tilde{I}^t\}_{t=1}^T$, while animating human hair in the photo,  where $T$ is  frame number and  $\Tilde{I}^t$ is a  frame of the generated video. However, animating hair in a still image is challenging, since the complexity of hair structure and dynamics poses new challenges compared with fluid element animation.

We address the above challenges by exploring two questions:  (1) What to animate in human hair? (2) How to automatically and naturally "blow"   hair?
Towards the first question, we observe that human visual system is much less sensitive to hair-strand-level motion than the wisp-level one in  real short videos, different from virtual/digitized environments. 
Motivated by this, we propose   a hair wisp extraction module to extract meaningful hair wisps  for still portrait photos  and   a wisp animation module that animates extracted hair wisps with  natural and pleasing motions.

\textbf{Overview.} 
We propose a framework for animating hair  for still portrait images.  Our  framework consists of three  steps, as shown in Fig. \ref{fig:framework}. First,  we propose an Instance-based Hair Wisp Extraction (IHWE) which automatically extracts hair units named \emph{hair wisps} that are locally grouped and would move consistently in a generated video, without relying on complex hair capture systems or user  assistance.
Second, we propose a hair wisp  animation module to animate hair wisps by predicting the spatiotemporal evolution of a hair wisp.
Third, with the animated hair wisps, we generate an  animated video by fusing animated hair wisps.

\subsection{Instance-based  Hair Wisp Extraction}
We propose to automatically extract hair wisps from a portrait photo for animating hair.    To the best of our knowledge, although some  hair capture approaches \cite{chai2016autohair,zhang2018modeling}  and face parsing approaches \cite{lin2019face,te2020edge,zheng2022decoupled,masi2020towards,li2020dual}  have deployed  hair segmentation algorithms for a single image, these approaches mainly focus on  extracting the whole hair region, rather than hair wisps. Due to the intricate  appearance and structure of hair, it is nontrivial to accurately segment the whole hair region \cite{chai2016autohair},  while  automatically extracting hair wisps  is much more challenging.  
For example, hair wisps are visually similar to each other in the same hair (see Fig. \ref{fig:teaser}),  which needs  extraction methods to discriminate  subtle  differences among them.

Different from existing work, we cast hair wisp extraction as an instance segmentation problem, inspired by the remarkable performance of  supervised  instance segmentation methods on  animals and humans. Thus, by treating a wisp  as an instance,  we can employ advanced instance segmentation networks to  extract hair wisps. Nevertheless, the difficulties lie in training data, especially the ground-truth annotations. It is  time-consuming and expensive to annotate each hair wisp in a real image manually. To address this challenge, we propose a training data construction scheme.  Below, we elaborate on the proposed instance-based wisp extraction module and  data construction scheme.

\textbf{Hair Wisp Extraction.}  Given a still portrait  image $I$, we present an  IHWE module   to predict instance masks $\mathbf{M}=\{M_i\}_{i=1}^N$ for hair wisps, where $N$ is the instance number,  $M_i\in R^{W\times H}$ is an instance mask of a hair wisp,  $W\times H$ is the size of image $I$.  
In particular, we first predict a matting map $\bar{M}$ which indicates the whole hair regions using \cite{chen2017rethinking}, to avoid irrelevant components negatively affecting instance segmentation results. With the matting map, we employ deep neural networks \cite{hu2021istr} to extract hair wisps, inspired by the success of supervised instance segmentation methods (\eg, \cite{hu2021istr} \cite{jain2022oneformer}). Thanks to  our instance segmentation,  hair wisps are adaptively extracted according to hair content, without  pre-defining wisp number $N$. After that, we further refine the shape of extracted hair wisps. 

   The issue is that these  networks  are task-oriented and rely on  specific training data.  However,  there are no proper   hair wisp datasets for instance segmentation. Although a few hair datasets (\cite{kim2021k,xiao2021sketchhairsalon}) have been presented, these datasets are constructed for extracting the whole hair region (\eg face parsing), which is not applicable to our task.

\textbf{Hair Wisp Dataset.}
We constructed a new hair dataset named Hair-Wisp dataset for instance segmentation, such that instance segmentation networks can be trained to detect hair wisps from portrait images in a supervised manner.   
However, it is laborious  to manually   annotate a segmentation mask for each wisp.  
To address this issue, we first propose a sketch-filled algorithm to generate ground-truth annotations of hair wisps for a portrait image.

Our sketch-filled algorithm generates ground-truth annotations of hair wisps inspired by \textit{guide} hairs used in hair modeling. In particular,  existing hair modeling methods \cite{plante2001layered,chang2002practical} generate/synthesize hair from a small number of guide hair strands that are representative to depict  hairstyles/structures. Inspired by this, we found a hair sketch dataset \cite{xiao2021sketchhairsalon} which provides user strokes that well indicate the distribution of hair wisps. Moreover, Xiao \etal \cite{xiao2021sketchhairsalon} synthesizes plausible hair using hair sketch. Here, given a portrait image in  hair sketch dataset, we exploit the associated hair sketch as a guided hair strand and annotate hair wisp by the sketch.

One way of generating wisp annotations  is to directly expand the hair sketch. However, the generated annotation results in such a manner  are low-quality. For example, the annotation results improperly  annotate parts of two neighboring wisps  as a wisp. In addition, the  annotated  regions of  hair wisps are rugged and lack smoothness.
{Instead, we design a  sketch-filled algorithm based on flood-fill \cite{levoy1981area} by expanding in the top, bottom, and right directions.} By ignoring the left-direction expansion, our algorithm well preserves the contour indicated by the strokes.

\begin{figure}[t]
\centering
\includegraphics[width=0.47\textwidth]{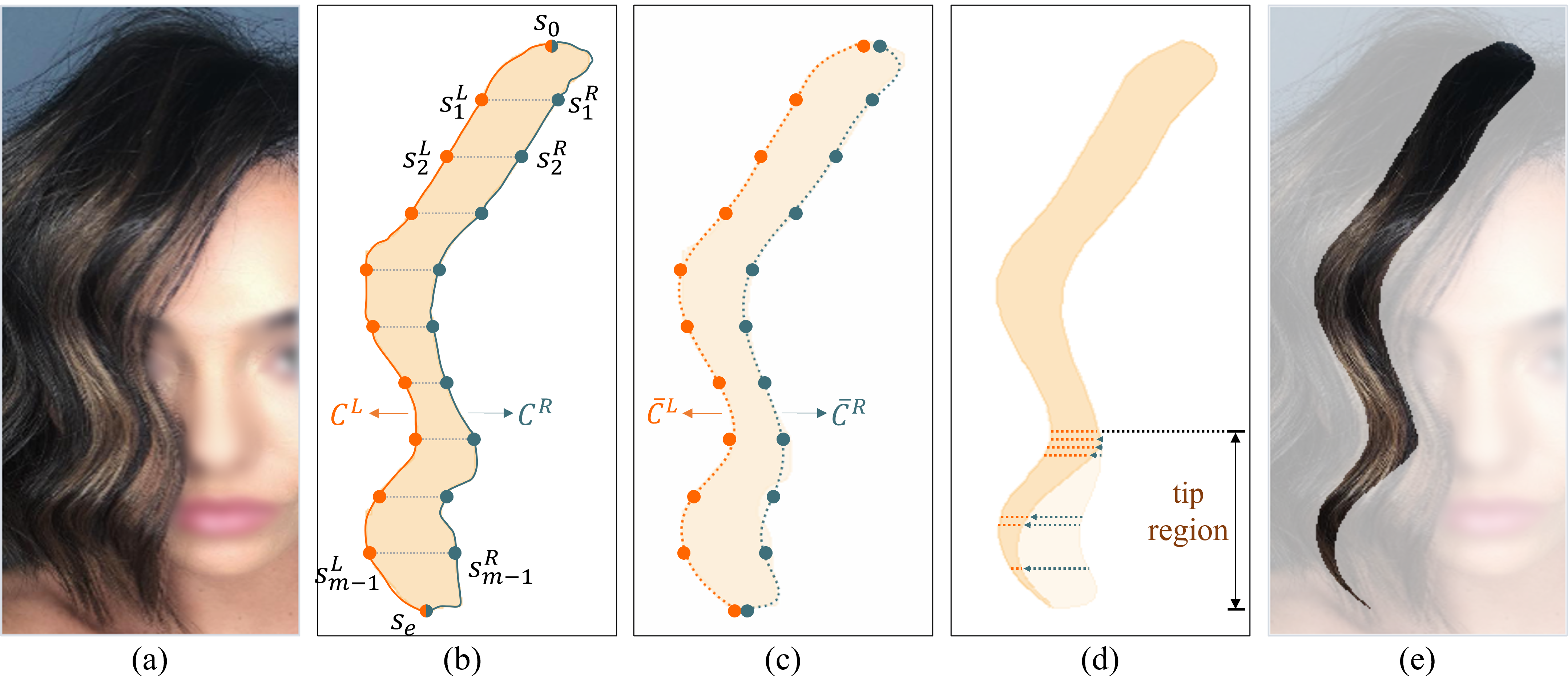}
\caption{Illustration of hair wisp shape refinement. Given a hair wisp in (b) extracted from an image (a) via instance segmentation, our method  smooths the left contour $C^L$ and the right one  $C^L$ in (b), and obtains refined   left/right contours ($\bar{C}^L$, $\bar{C}^R$) in (c). We then sharpen the tip region of the wisp in (d).    }
\label{fig:post-process}
\end{figure}

 With the above annotation generation, we construct  the Hair-Wisp dataset from the hair sketch dataset  \cite{xiao2021sketchhairsalon}. Our dataset  includes 4500   images and ground-truth annotations of hair wisps, which covers  challenging cases such as   braid hairstyles. It is worth noting that our method does need completely accurate annotations to train hair wisp extraction networks, as we design shape refinement algorithms below. Please refer to the supplementary for more details.

 \textbf{Hair Wisp Shape Refinement.} After exploiting instance segmentation networks to extract hair wisps from the input photo, we further refine the extracted hair wisps.   
 Due to the limitation of instance segmentation networks, the contours of  an extracted hair wisp is not smooth and its tip is coarse. (see Fig.~\ref{fig:post-process} (a)). Hence, we  smooth the contours of extracted hair wisps and then sharpen  wisp tips. 
Given an extracted wisp, we first split its contour into left and right contours using  its  top-most and bottom-most points. The left/right contour of extracted hair is uniformly divided to obtain sample points 
(see Fig. \ref{fig:post-process} (b)).  Let $\{s_0, s^z_1, ...,  s_e\}$ be sample points on $z$ contour $C^z$, where $z=L$ denote left and $z=R$  right. The left/right contour is smoothed by polynomial regression $g$:
\begin{equation}
\bar {C}^z= g(s_0, s_1^z, ...,  s_e)
\end{equation}
We then sharpen the tip region of a hair wisp by linearly shrinking the width of the tip region (see Fig.~\ref{fig:post-process} (d)).

\subsection{Hair Wisp Animation}

We describe how to  generate  motions that animate human hair wisps in a still image.  Recent single-image-to-video methods \cite{mahapatra2021controllable,Holynski_2021_CVPR} use deep learning techniques to animate fluid elements, since fluid motion can be approximated via a static velocity field. However, since hair motions are  complex \cite{wang2022neuwigs} and the input is only a single image, it is non-trivial to create hair motions using  deep learning techniques. Instead, we animate hair wisps based on physical models.

There are two new challenges posed by generating hair wisp motions.  First,  the dynamics of a  hair wisp not only include motion displacement but also the shape deformation of the wisp.    As a result,   if we  apply physically based animation algorithms (\eg simulation \cite{selle2008mass,daviet2011hybrid}) to  hair wisps  by  treating a hair wisp as a strand,  the algorithm fails to model shape deformation, leading to unrealistic  motions and artifacts.
Second, the hair root regions  of  many extracted hair wisps are partially occluded by other objects such as  the face in a portrait photo.  However, the motion of  a hair wisp is continuous and starts at the scalp of the head. In other words, given a wisp, its local region  closer to the head scalp affects the motions of other farther regions in it. Hence, it is difficult to create motions for partially-occluded  hair wisps.

We address these challenges by  proposing a  hair wisp animation module. In particular, 
to model shape deformation and motion displacement, we represent hair wisps with meshes.
We then impose mass-spring systems on meshes to temporally drive the hair wisps for animation, since mass-spring systems have been used to simulate various non-rigid objects such as cloth,  providing a simple yet effective solution.
Furthermore,  we recognize partially-occluded hair wisps that are not connected to the head scalp, and explicitly approximate motions for these wisps. 

\textbf{Wisp classification.} We classify hair wisps into two categories: scalp-connected and  scalp-unconnected. In particular, we first obtain  the forehead contour $C$ of the human face in input image $I$ by detecting  facial keypoints along the forehead contour and  connecting them. If a hair wisp  intersects with the forehead contour, it is classified as  scalp-wisp; otherwise non-scalp-wisp.

\textbf{Multi-layer Mesh Representation.}
Given an input image $I$, we represent hair wisps in $I$ by  multiple layers of meshes, rather than a single mesh. In particular,   the $k$-th  hair wisp  $W^{k}$    is represented by  a triangle mesh $\{\mathbf{X}^k,\mathbf{E}^k\}$, where  $\mathbf{E}^k$ is the set of triangle edges, $\mathbf{X}^k=\{\mathbf{x}^k_i\}$ is the set of mesh vertexes and $\mathbf{x}^k_i\in \mathbb{R}^2$  is the postion of the $i$-th vertex.  We construct a  mesh  by Delaunay Triangulation. To improve the freedom of degree (FOD) of a mesh, we increase  the number of vertexes in a mesh by dividing circumscribed rectangles of the hair wisps into 6 $\times$ 6 grids.

Based on the above mesh representation,  predicting the dynamics of  a  hair wisp  $W^{k}$ is  to predict the spatial-temporal evolution of its mesh along the time axis.
 Let $\mathbf{\Tilde{X}}^{k,t}$  be a set including  positions of vertexes  $\mathbf{X}^k$  at time $t$. After obtaining  $\mathbf{\Tilde{X}}^{k,t}$, we can generate the  animated hair wisp  $\Tilde{W}^{k,t}$ for a generated frame $\Tilde{I}^t$  by   warping   $W^{k}$:

\begin{equation}
 \Tilde{W}^{k,t} = \varpi (W^{k}, \mathbf{X}, \mathbf{\Tilde{X}}^{k,t}),
\end{equation}
where $\varpi$ is the warping operator. Here, we employ the thin plate spline algorithm \cite{bookstein1989principal} to obtain dense warping fields.

 \begin{figure}[t]
\centering

\includegraphics[width=0.48\textwidth]{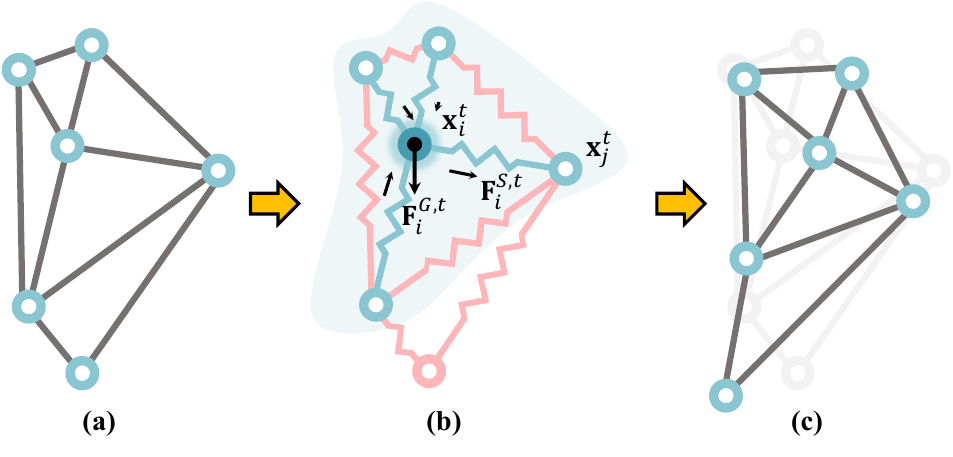}
\caption{Given a mesh of a hair wisp (a),  springs that are imposed to  connect neighboring vertexes $\mathbf{x}^t$ in (b) drive the mesh by  forces $F$, such that warped mesh (c) approximates the spatiotemporal evolution of the wisp.  }
\label{fig:spring}
\end{figure}

\textbf{Wisp Motion Prediction.}  To estimate vertex positions  $\mathbf{\Tilde{X}}^{k,t}$ at time $t$ and simulate motions for hair wisp $W^k$, we impose mass-spring systems on the  mesh of $W^k$ .   In particular, we directly use  a mesh vertex as a  particle with mass $m$, and each two neighboring particles are connected by  a spring (see Fig. \ref{fig:spring}). Consequently, the motion of a vertex is affected by spring forces and other forces. Here, we only consider gravity and spring forces for the sake of simplicity. Below, We omit the superscript $k$ for brevity.  The accumulated force $\mathbf{F}^{t}$ that acts on vertex $x_i^t\in \mathbf{\Tilde{X}}^{k,t}$ is measured as the gravity force $\mathbf{F}_i^{G,t}$ and spring force $\mathbf{F}_i^{S,t} $:

\begin{equation}
\mathbf{F}^{t}_i= \mathbf{F}_i^{G,t} + \mathbf{F}_i^{S,t} 
\end{equation}

The  spring force $\mathbf{F}_i^{S,t}$ is computed   according to Hooke's law. 
Since a vertex is connected by multiple springs, we compute $\mathbf{F}_i^{S,t}$  from  all springs connected to $\mathbf{x}_i^{t}$ :
\begin{equation}
 \mathbf{F}_i^{S,t}  = \sum_{j \in \mathcal{N}} K\cdot ( \mathbf{x}_i^{t}  -\mathbf{x}_j^{t}- |\mathbf{e}_{ij}|)
\end{equation}
where $K$ is the spring constant indicating force strength, $\mathbf{e}_{ij}$ is a spring connected to the $i$-th vertex $\mathbf{x}_i^{t}$, $|\mathbf{e}_{ij}|$ is the spring length, $\mathcal{N}$ is the set including the indexes of neighboring vertexes of $\mathbf{x}_i^{t}$. 

With the accumulated force, the vertex would be moving. We thereby predict  vertex positions at time $t$ by estimating acceleration and velocity. In particular, according to Newton's second law of motion,  the acceleration of a vertex at time $t$ is calculated: 
\begin{equation}
    a^t= \frac{\mathrm{d}\mathbf{v}_i(t)}{\mathrm{d}t}=\frac{\mathbf{F}^{t}_i}{m}
\label{eq:acc}
\end{equation}
where $v^{t}_i$ is  the velocity  of vertex $\mathbf{x}_i^{t}$.
The position of the mass meets ODE: $\mathbf{v}_i(t)= \frac{\mathrm{d}\mathbf{x}_i(t)}{\mathrm{d}t}$.
By using  Euler's method in  Eq. \ref{eq:acc},  we predict vertex positions at the time axis as:
\begin{align}
&\mathbf{v}_i(t+\Delta t)= \mathbf{v}_i(t) + a^t_i\cdot \Delta t\\
 &   \mathbf{x}_i(t+\Delta t)=\mathbf{x}_i(t)+ \mathbf{v}_i(t+\Delta t)\cdot \Delta t
\end{align}
where initial velocity $\mathbf{v}_i(0)$ is preset.  We can set $\mathbf{v}_i(0)=[0\;0]$ or set     $\mathbf{v}_i(0)$ to be a non-zero vector to indicate wind force.  In addition, since vertexes on the head scalp  are  static,  the positions of these vertexes are fixed and are equal to the original values.

We can not directly  apply the above wisp motion model to unconnected-scalp wisps.   Since no vertex of an unconnected-scalp wisp is fixed,  gravity would pull down all vertexes in the wisp, forcing the wisp to  keep falling.  We address the issue by approximating  motions  for  the topmost vertex of an unconnected-scalp wisp. In particular, we  construct a  mesh on the entire hair, where the mesh is named auxiliary mesh. We then use  the wisp motion model to predict the motion of the auxiliary mesh, while fixing the positions of vertexes lying on the head scalp. For  the topmost vertex of  an unconnected-scalp wisp,  we   approximate its motion by   that of  the auxiliary mesh's vertex which has the same position as it. By such approximation,   the topmost vertex of  an unconnected-scalp wisp is relatively fixed to  the auxiliary mesh. We apply the wisp motion model to the remaining vertexes of an unconnected-scalp wisp.

Note that existing physically-based  hair animation methods (\eg \cite{chai2013dynamic}) are mainly designed for reconstructed hair stands or virtual humans. These  methods first simulate motions for a small number of guide hair  strands, and then approximate motions for the rest strands  from that of the guide ones.  However, it is extremely difficult to accurately extract guide  strands without user assistance (\eg \cite{chai2013dynamic}), given only a still image. Differently, our method  generates motions for hair wisps and automatically  detects hair wisps, which largely reduces the computational difficulty and is  simpler to apply without user assistance.

\subsection{Video Generation} 
We describe the progress of generating a video from  the extracted hair wisps and  their predicted dynamics. 
 
 \textbf{Depth-aware frame composition.} A naive way of rendering a frame $\Tilde{I}^t$  is to composite the background and the warped versions $ \{\Tilde{W}^{1,t}, ...,\Tilde{W}^{k,t},...\}$ of all extracted hair wisp, where the background is extracted by removing hair regions from original image $I$. However, this would lead to improper occlusion relationship between wisps and face. For example, after the composition,  some warped hair wisps that should be behind the face  may  occlude the face, which often leads to unrealistic motions. To address this issue,  we introduce  depth information to guide the composition \cite{li2018depth,li2021high}. In particular, we use a face parsing algorithm \cite{chen2017rethinking} to extract face regions. By  applying a depth estimation algorithm \cite{ranftl2021vision} to image $I$, we   obtain the  relative depth relationship  between the face and a hair wisp.  
The hair wisp layers and background are sorted by the depth and the height of hair wisps. 
 We then  composite the sorted  warped wisps  and the  background to render frame $\Tilde{I}^t$, following Painter’s algorithm.
 In addition, our method refines the background layer by inpainting \cite{Suvorov_2022_WACV} missing regions, which avoids hole-filling for each generated frame.

\begin{figure*}[t]
	\centering	
	\begin{minipage}[t]{0.195\textwidth}
		\centering

  \includegraphics[bb=0 0 512 770,width=1\textwidth]{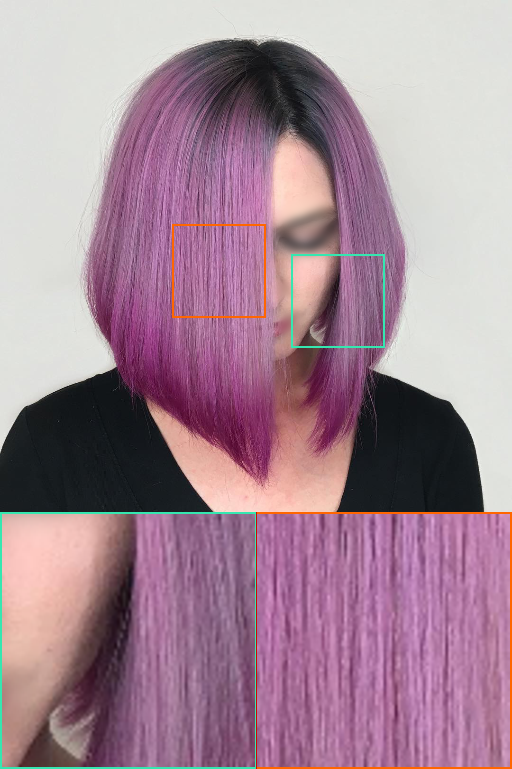}
 \includegraphics[bb=0 0 564 848,width=1\textwidth]{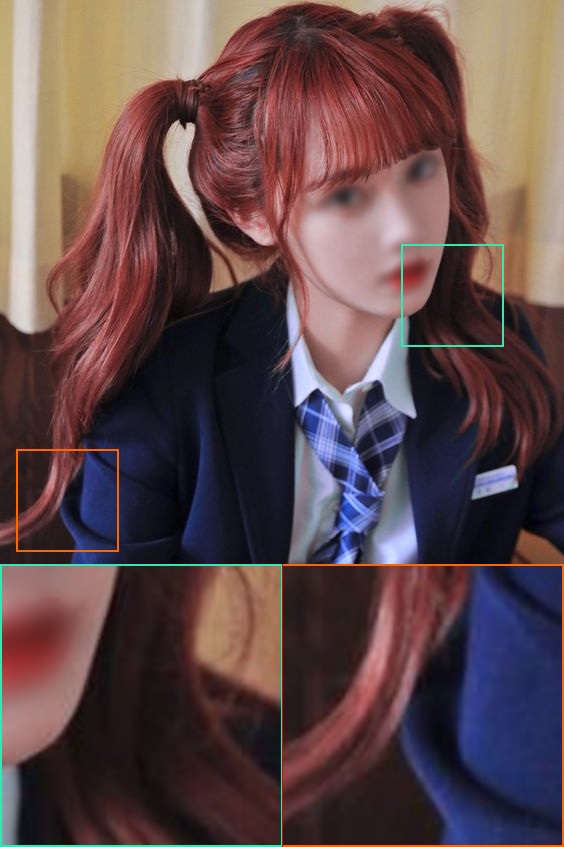}

		\footnotesize 	Input image 
	\end{minipage}
	\begin{minipage}[t]{0.195\textwidth}
		\centering

  \includegraphics[bb=0 0 512 770,width=1\textwidth]{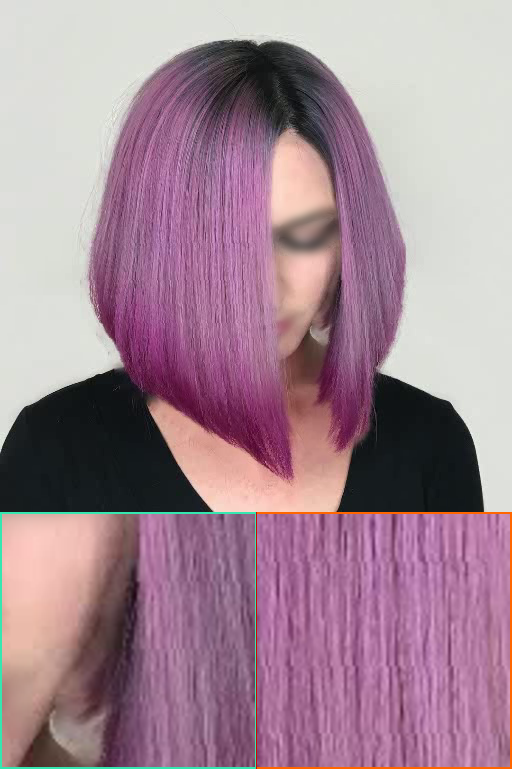}
 \includegraphics[bb=0 0 564 848,width=1\textwidth]{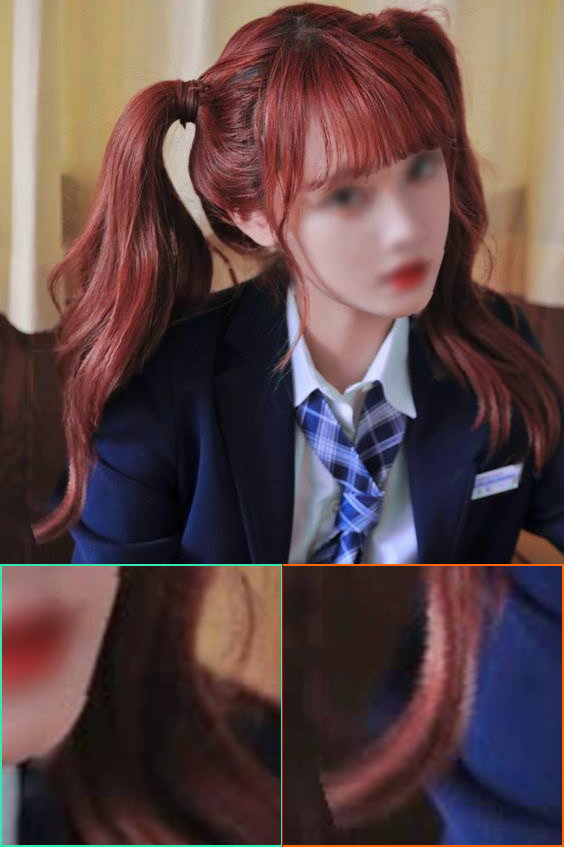}

		\footnotesize Chuang \etal \cite{chuang2005animating}
	\end{minipage}
	\begin{minipage}[t]{0.195\textwidth}
		\centering

  \includegraphics[bb=0 0 512 770,width=1\textwidth]{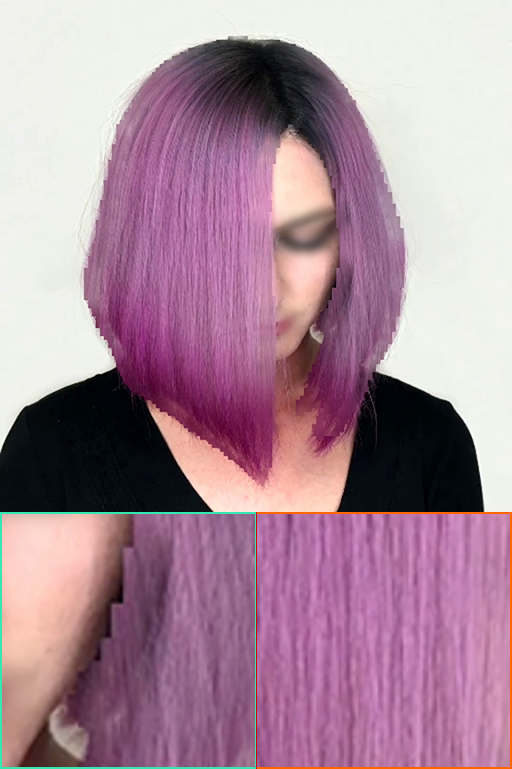}
 \includegraphics[bb=0 0 564 848,width=1\textwidth]{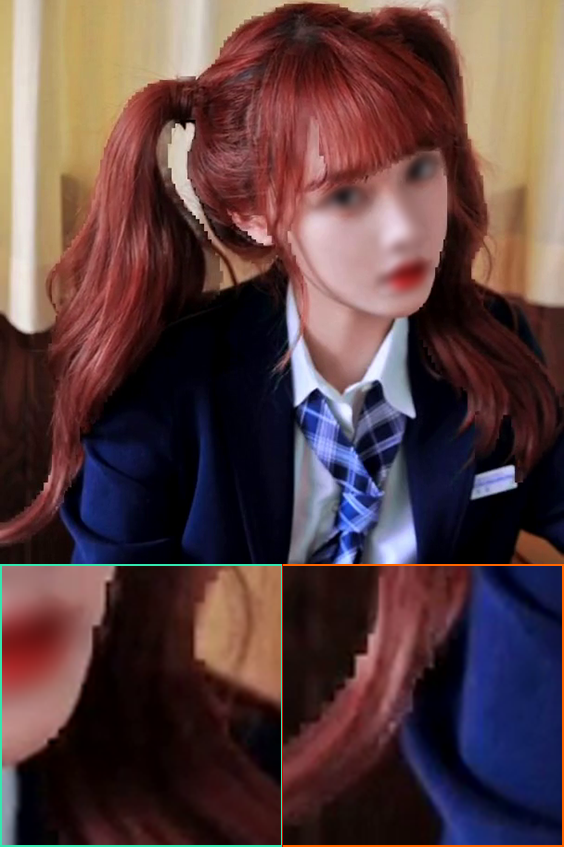}

		\footnotesize 	Halperin  \etal \cite{halperin2021endless}
	\end{minipage}
	\begin{minipage}[t]{0.195\textwidth}
		\centering

  \includegraphics[bb=0 0 512 770,width=1\textwidth]{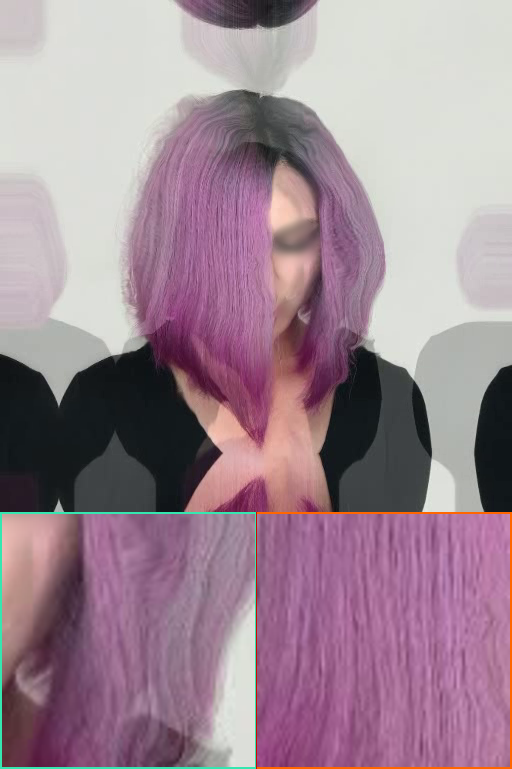}
  \includegraphics[bb=0 0 564 848,width=1\textwidth]{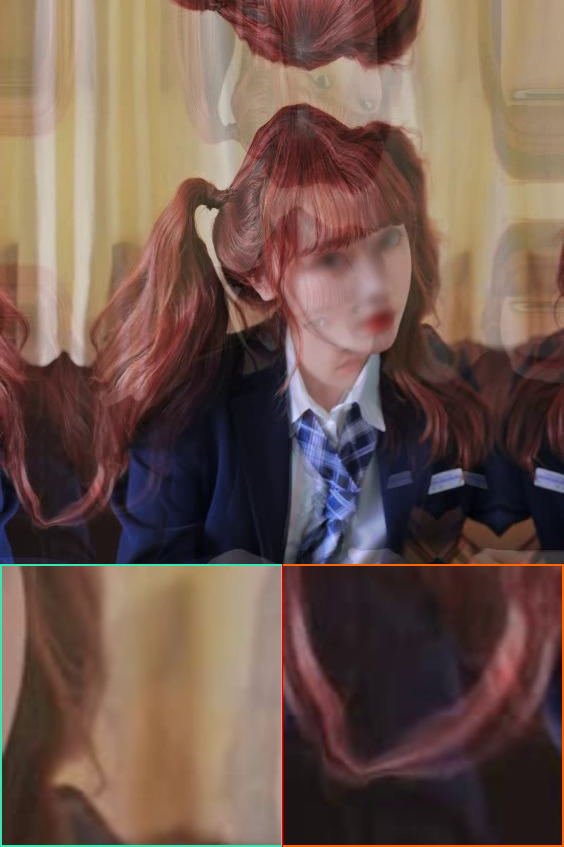}

		\footnotesize 	Endo \etal \cite{endoSA2019}
	\end{minipage}
	\begin{minipage}[t]{0.195\textwidth}
		\centering

  \includegraphics[bb=0 0  512 770,width=1\textwidth]{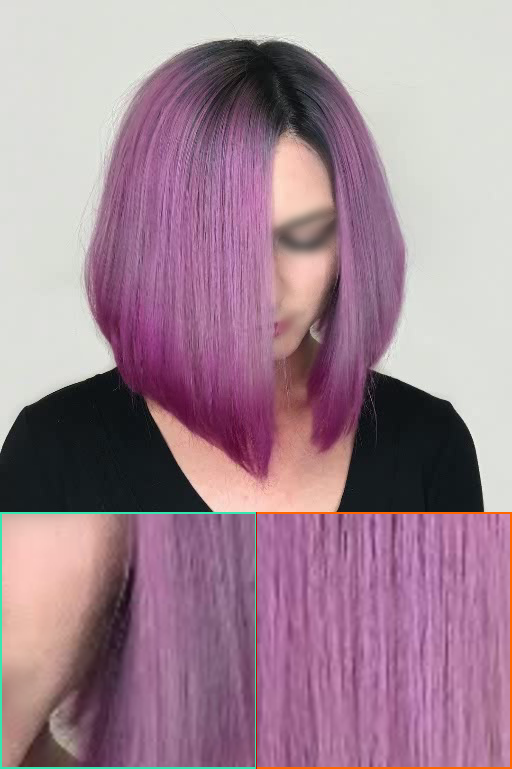}
 \includegraphics[bb=0 0  564 848,width=1\textwidth]{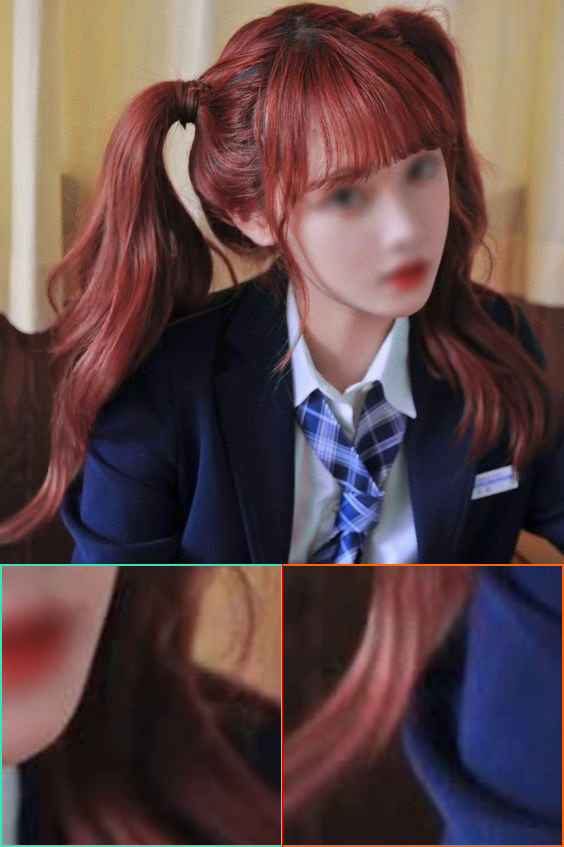}

		\footnotesize Ours
	\end{minipage}
	
    	\caption{Qualitative comparisons of our approach with single-image-to-video generation  methods \etal \cite{chuang2005animating} \cite{halperin2021endless} \cite{endoSA2019}. In each image, we show a full frame at the top  and zoom-in  rectangle regions marked by red/green at the bottom. 
	}
	\label{fig: compared_result1}
\end{figure*}

\section{Experiments}

\subsection{Baselines and Implementation Details}
 We compare our method  with three state-of-the-art methods in single-image-to-video generation: Halperin  \etal \cite{halperin2021endless}, Endo \etal \cite{endoSA2019} and 
Chuang \etal \cite{chuang2005animating}\footnote{To the best of our knowledge, most hair animation methods  are designed for synthetic data (\eg games and virtual reality). These methods are not applicable to a real portrait photo, since they  require hair strand information which is significantly difficult to acquire without complex hair capture systems.}, where   Halperin  \etal and Endo \etal  are recent work using deep learning techniques. 
Both   Endo \etal  and  Chuang \etal require user assistance, where Chuang \etal needs users to manually decompose the input image into layers, and Halperin  \etal requires a  user-provided mask indicating which regions are moving and a general motion direction. Therefore, we manually specify motion directions and mask hair regions   for  Chuang \etal and Halperin  \etal in
 our experiments. 
To implement Endo \etal, we  train its networks   on real cinemagraphs containing  hair blowing. More implementation details and comparisons to other competitors  are provided in supplementary materials, where we additionally show the results of applying our method to anime images and clothes.  Portrait photoes used in our experiments are from dataset SketchHairSalon \cite{xiao2021sketchhairsalon}.

\subsection{Evaluation Metrics}
We use two evaluation metrics to assess the performance of our method on video quality and temporal consistency. 

\textbf{Frechet Video Distance} (FVD)\cite{unterthiner2018towards}. FVD is a standard metric that has been popularly used for evaluating the quality of generated videos in recent video synthesis work (\eg \cite{mahapatra2021controllable, Dorkenwald_2021_CVPR}).  FVD assesses the quality of generated videos by measuring the data distribution between generated and real videos, based on FID \cite{heusel2017gans}. 
The pre-trained I3D networks are employed \cite{szegedy2016rethinking} to extract video features, where the networks are trained on  Kinetics dataset \cite{kay2017kinetics}. 
A lower FVD score indicates a higher quality of  generated videos.

\begin{figure*}[!hptb]
	\centering	
	\begin{minipage}[t]{0.195\textwidth}
		\centering
  \includegraphics[bb=0 0 302 454,width=1\textwidth]{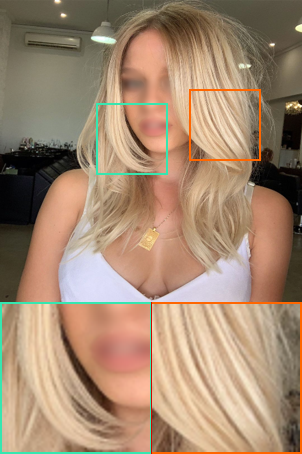}
		\footnotesize Input
	\end{minipage}
     \begin{minipage}[t]{0.195\textwidth}
		\centering
  \includegraphics[bb=0 0 302 454,width=1\textwidth]{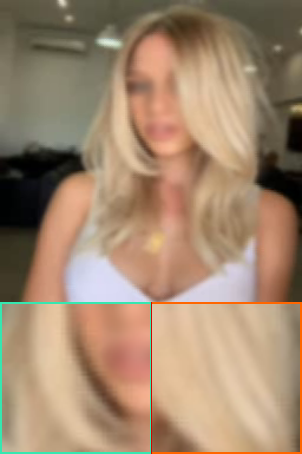}
		\footnotesize Hu \etal ~\cite{hu2023dmvfn}
	\end{minipage}
	\begin{minipage}[t]{0.195\textwidth}
		\centering
  \includegraphics[bb=0 0 302 454,width=1\textwidth]{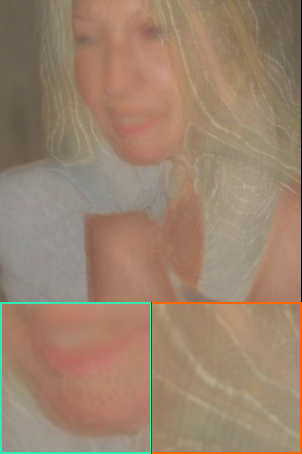}
		\footnotesize 	Qiu \etal ~\cite{StyleFaceV} 
	\end{minipage}
 \begin{minipage}[t]{0.195\textwidth}
		\centering
  \includegraphics[bb=0 0 302 454,width=1\textwidth]{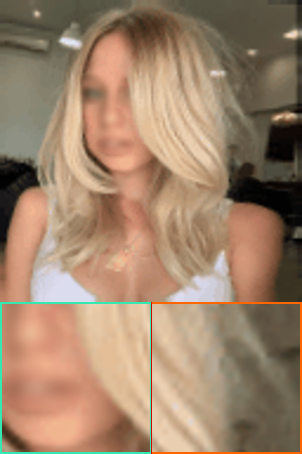}
		\footnotesize 	Ni \etal ~\cite{ni2023conditional} 
	\end{minipage}
		\begin{minipage}[t]{0.195\textwidth}
		\centering
  \includegraphics[bb=0 0 302 454,width=1\textwidth]{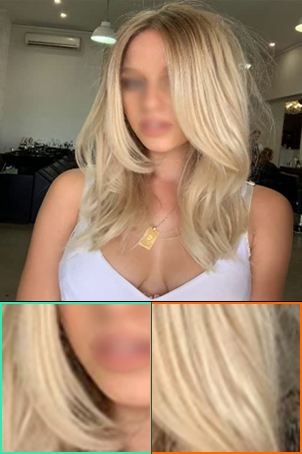}
		\footnotesize 	Ours
	\end{minipage}

 \caption{Qualitative Comparison results of our approach with recent video prediction method~\cite{hu2023dmvfn}, GAN-based method ~\cite{StyleFaceV}, diffusion model ~\cite{ni2023conditional}.  In each image, we  zoom in  rectangle regions marked by red/green at the bottom.} 
	\label{fig: qual_comp_SOTAs}
\end{figure*}

\textbf{Warping Error} $E_{warp}$ \cite{lei2020blind}. We adopt $E_{warp}$ to measure the temporal inconsistency of generated videos, following \cite{lei2022deep,dai2022video}.  To evaluate short- and long-term consistency of a video, $E_{warp}$  computes the warping errors between consecutive frames as well as that of  each frame  and the first frame. A lower $E_{warp}$  value indicates better temporal consistency.

\subsection{Comparison}

\textbf{Quantitative and Qualitative Results.} 
Tab. \ref{tab:comparison} shows our method achieves the best FVD and $E_{warp}$ value, indicating our method outperforms the three state-of-the-art methods in terms of  video quality and temporal consistency.
The supplementary video and Figs. \ref{fig: compared_result1} and \ref{fig: qual_comp_SOTAs}  demonstrate qualitative  comparison results,  showing our approach outperforms state-of-the-art methods on various testing images. More comparisons are provided in supplementary materials.
\begin{table}[t]

\centering

\begin{tabular}{l|cc}
\toprule  
Metric & FVD  $\downarrow$   & $E_{warp}$ $\downarrow$ \\
\midrule 
 Chuang \etal~\cite{chuang2005animating} & 1263.44 & 547.64   \\
 Halperin \etal~\cite{halperin2021endless}  & 1778.24 & 661.34   \\
 Endo \etal~\cite{endoSA2019} & 2026.59 & 1280.66\\
      Qiu \etal~\cite{StyleFaceV} & 1329.46 & 573.78 \\
          Hu \etal~\cite{hu2023dmvfn} & 1301.96 & 969.96 \\
      Ni \etal~\cite{ni2023conditional} & 1192.75 & \textbf{319.05}\\
 Ours & \textbf{1153.98}& 521.96\\ 
\bottomrule 
\end{tabular}
 \caption{Quantitative comparison results on 114 portrait images of SketchHairSalon\cite{xiao2021sketchhairsalon}. Our method outperforms state-of-the-art approaches. }
\label{tab:comparison}
\end{table}

 Endo \etal  leads to noticeable distortions in Fig. \ref{fig: compared_result1} and  supplementary video. This is because  Endo \etal  employ networks to learn motion fields from training data,  while it is difficult to directly train the networks that can effectively  capture hair dynamics  due to the complex motion space of hair dynamics.
 Halperin  \etal  combines a conditional random field with a local feature descriptor to compute a temporally and spatially continuous displacement field,    which achieves remarkable animation performance for objects with periodic structures. However,  since such  a displacement field improperly drives  all hair regions to consistently move in the same direction,  Halperin  \etal  create  unrealistic and unnatural motions for hair (see the supplementary video). 
In addition, Halperin  \etal   introduces   jagged artifacts into the boundary of  the face and hair (see  Fig. \ref{fig: compared_result1}). 
Chuang \etal  often leads to jitter artifacts in animated hairs, since this method approximates motion as harmonic oscillations, which is more suitable for branches and grass rather than hair. 
 In contrast,  our instance-based hair wisp extraction enables our method to generate various motions among wisps through extracting hair wisps,  which is helpful to simulate complex and natural motions.  Moreover, our hair wisp animation module effectively generates realistic motions for a hair wisp by building a physics-based  model.
 As a result,  our method generates videos with the highest quality compared with all baseline methods.

In addition, Fig. \ref{fig: qual_comp_SOTAs} and Tab.~\ref{tab:comparison} compares our method with the state-of-the-art  video prediction method~\cite{hu2023dmvfn}, GAN-based method ~\cite{StyleFaceV}, and diffusion model ~\cite{ni2023conditional}. They are the most recent  single-image-to-video generation methods. As shown in  Fig. \ref{fig: qual_comp_SOTAs} and Tab.~\ref{tab:comparison}, Qiu \etal~\cite{StyleFaceV} introduces  noticeable distortions, while Hu \etal~\cite{hu2023dmvfn} creates unnatural motions,  due to the challenges of hair animation. Similarly,  Ni \etal \cite{ni2023conditional}  does not properly model  complicated hair dynamics, leading to static scenes and introducing distortions in the left chin in Fig.~\ref{fig: qual_comp_SOTAs}.
In contrast, our method achieves the best performance in Fig. \ref{fig: qual_comp_SOTAs} and Tab.~\ref{tab:comparison}, compared with these methods.

\begin{figure}[!htbp]
	\centering	
	\begin{minipage}[t]{0.155\textwidth}
		\centering
  \includegraphics[bb=0 0 512 770,width=1\textwidth]{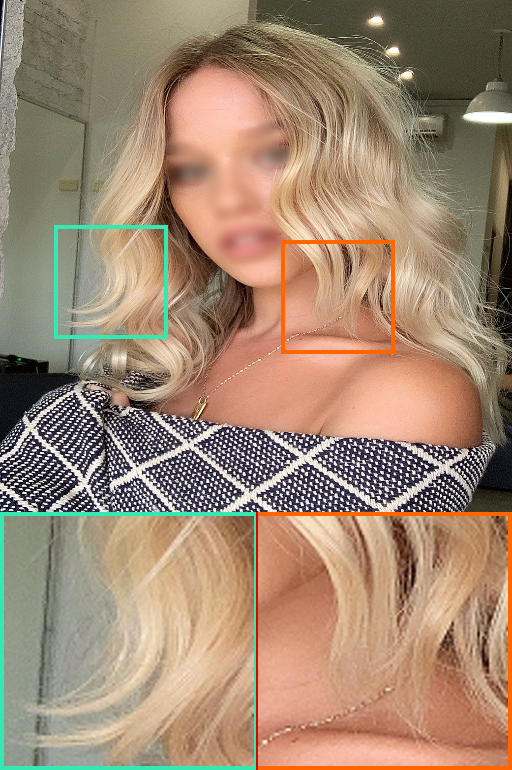}
		\footnotesize Original
	\end{minipage}
	\begin{minipage}[t]{0.155\textwidth}
		\centering
  \includegraphics[bb=0 0 512 770,width=1\textwidth]{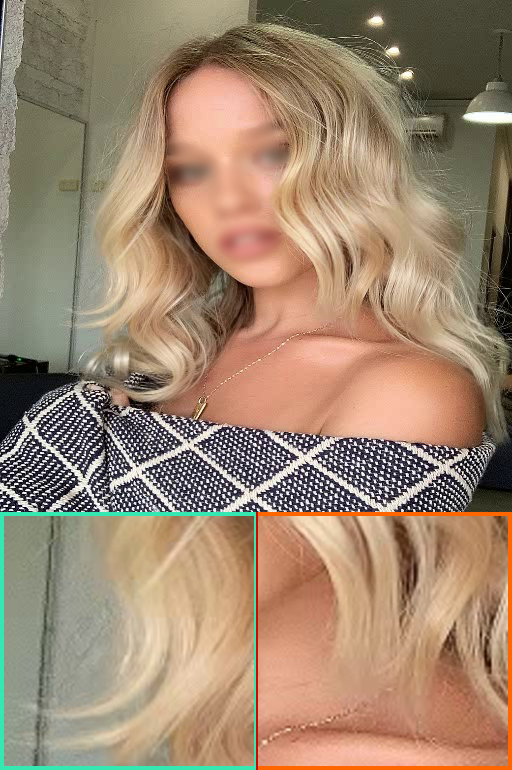}
		\footnotesize 	w/o IHWE w WE
	\end{minipage}
	\begin{minipage}[t]{0.155\textwidth}
		\centering
  \includegraphics[bb=0 0 512 770,width=1\textwidth]{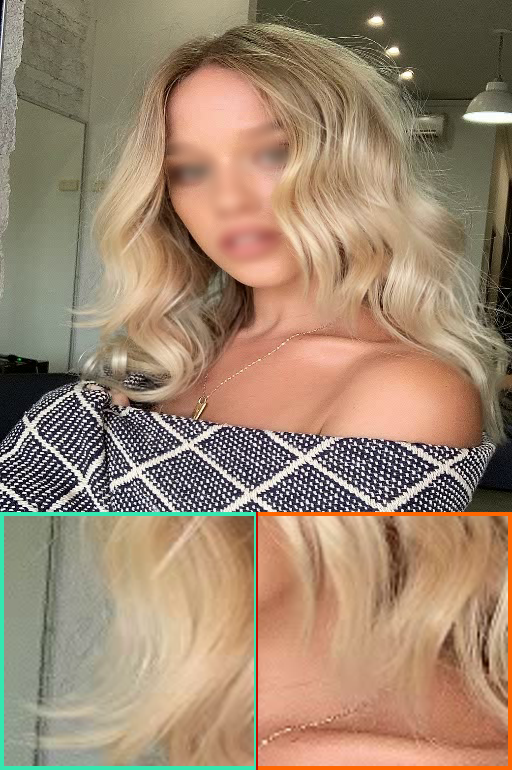}
		\footnotesize 	Ours
	\end{minipage}
	
	\caption{Effect of hair wisp extraction} 

	\label{fig: ablation_wholehair}
\end{figure}

\textbf{User Study.}
We conduct a subjective user study to evaluate our method. 18 subjects of various ages  are invited to participate in our user study.  Following \cite{huang2018multimodal,wang2018high}, we adopt  paired comparison which is widely used to  subjectively evaluate image/video quality of generated images/videos.
For each subject, we display a still  photo and two animated videos generated by different methods, where the photo is in the center and two videos are randomly presented side-by-side.   All subjects have no prior knowledge of technical details.  We asked each subject a question:  \textit{which video do you prefer by considering the original image?}

We compare  our method with  Halperin  \etal \cite{halperin2021endless}, Endo \etal \cite{endoSA2019} and 
Chuang \etal \cite{chuang2005animating} on 10 testing photos. 
In total, 81.5\% of subjects are in favor of our method.  These results indicate our method generates the most pleasing video, compared with all these state-of-the-art methods,     although testing data contain complex backgrounds and  various hairstyles and head poses.

\begin{figure}[t]
	\centering	
	\begin{minipage}[t]{0.155\textwidth}
		\centering
  \includegraphics[bb=0 0 512 770,width=1\textwidth]{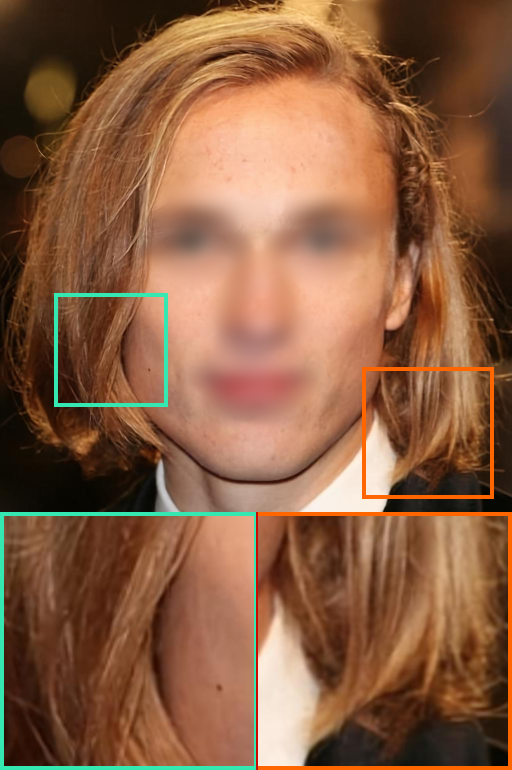}
		\footnotesize Original
	\end{minipage}
	\begin{minipage}[t]{0.155\textwidth}
		\centering
  \includegraphics[bb=0 0 512 770,width=1\textwidth]{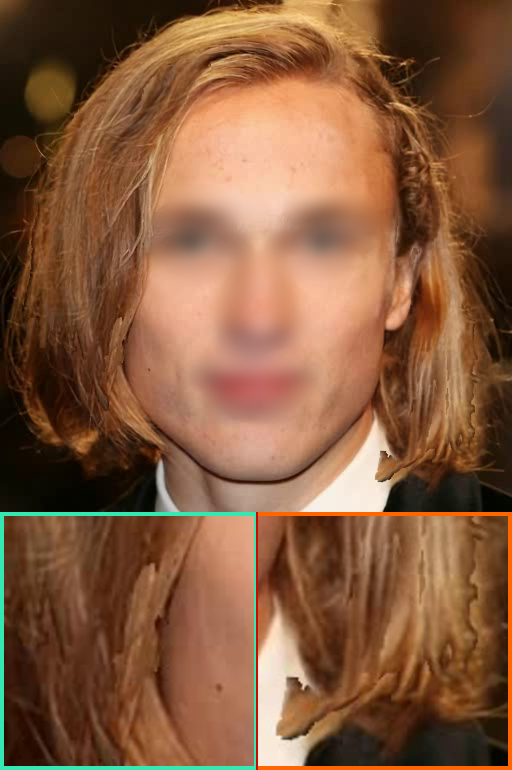}
		\footnotesize  w/o IHWE w FF
	\end{minipage}
	\begin{minipage}[t]{0.155\textwidth}
		\centering
  \includegraphics[bb=0 0 512 770,width=1\textwidth]{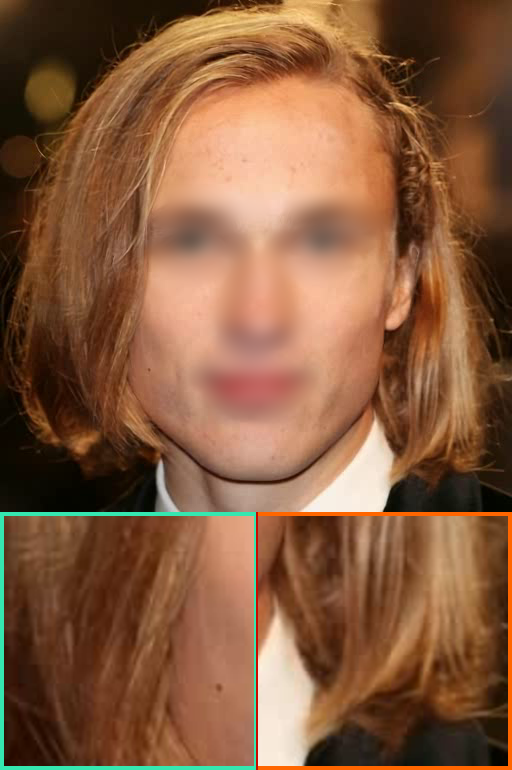}
		\footnotesize 	Ours
	\end{minipage}
	
	\caption{Visual comparison of the contributions of our instance-based hair wisp extraction module}

	\label{fig: ablation_flood}
\end{figure}

\subsection{Ablation Studies}

\textbf{Instance-based Hair Wisp Extraction} (IHWE.) We first validate the importance of hair wisp extraction in animating hair in a real image.  We build a baseline named \textbf{w/o IHWE w WE}  which removes our IHWE from our method and  extracts the entire hair region instead. Fig. \ref{fig: ablation_wholehair} shows \textbf{w/o IHWE w WE} creates unrealistic motions,  which  improperly enforces all hair wisps to undergo spatially consistent   movements and fails to model the complexity of hair dynamics. Instead, thanks to our hair wisp extraction, our method well creates variations among   hair wisps (see supplemental videos). 

 We then show  the effectiveness of IHWE by the second of baseline named \textbf{w/o IHWE w FF}.  \textbf{w/o IHWE w FF} employs a Flood Fill algorithm to extract hair wisps, instead of using  IHWE. Tab. \ref{tab: abaltion} shows  \textbf{w/o IHWE w FF} generates videos with the lowest quality (FVD) and worst temporal consistency ($E_{warp}$).  Fig. \ref{fig: ablation_flood} and supplemental videos show \textbf{w/o IHWE w FF} tears many hair wisps, leading to noticeable discontinuity artifacts within a hair wisp.

\begin{figure}[t]
	\centering	
	\begin{minipage}[t]{0.155\textwidth}
		\centering
  \includegraphics[bb=0 0 512 770,width=1\textwidth]{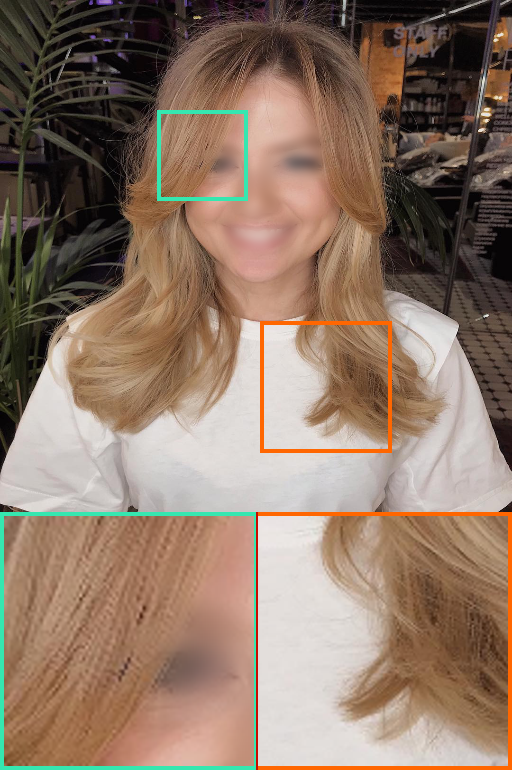}
		\footnotesize Original
	\end{minipage}
	\begin{minipage}[t]{0.155\textwidth}
		\centering
  \includegraphics[bb=0 0 512 770,width=1\textwidth]{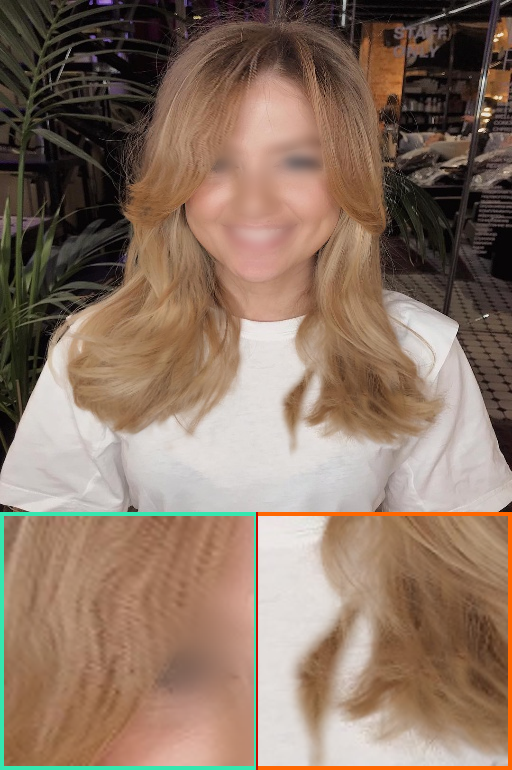}
		\footnotesize 	 w/o HWA
	\end{minipage}
	\begin{minipage}[t]{0.155\textwidth}
		\centering
  \includegraphics[bb=0 0 512 770,width=1\textwidth]{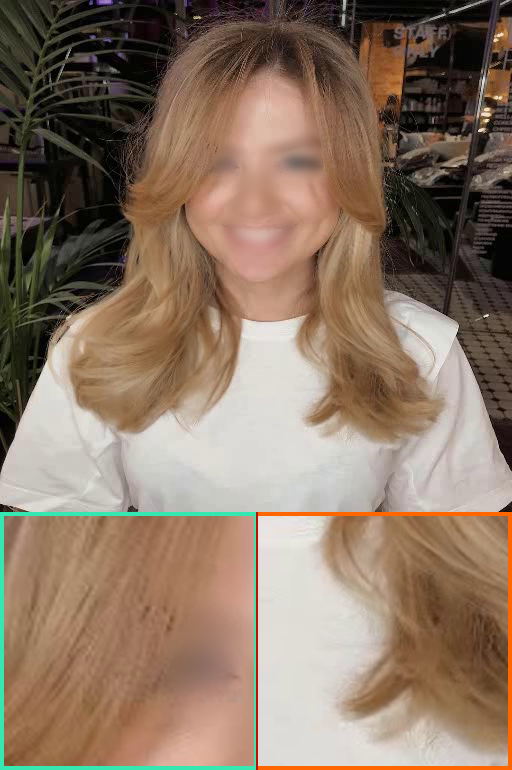}
		\footnotesize 	Ours
	\end{minipage}
	
	\caption{Visual comparison of the contributions of hair wisp animation module. w/o HWA results in distortions (region masked by cyan) and unrealistic motions (region masked by orange).}

	\label{fig: abla_stand_motion}
\end{figure}

\textbf{Hair Wisp Animation (HWA).} To evaluate   the effectiveness of our hair wisp animation, we build a baseline named \textbf{w/o HWA} by treating a hair wisp as a strand and driving it by a mass-spring system, like synthetic-hair-based  animation. \textbf{w/o HWA} achieves  a worse FVD value than our method in Tab. \ref{tab: abaltion}, since     such animation is unable to delicately model the inner dynamics of hair wisps (\eg temporal shape deformation of a hair wisp).
Fig. \ref {fig: abla_stand_motion} also  shows  \textbf{w/o HWA}    introduces  flickering artifacts.

\begin{table}[!t]
\centering
\begin{tabular}{l|cc}
\toprule  
Method & FVD  $\downarrow$  &  $E_{warp}$ $\downarrow$   \\
\midrule 
w/o IHWE w WE  & 1439.97 & 575.60 \\
w/o IHWE w FF & 1591.90 & 647.59 \\
 w/o HWA & 1367.24 & 586.40 \\
Ours & \textbf{1153.99} & \textbf{521.96}\\
\bottomrule 
\end{tabular}
\caption{The ablation study results.  Best in bold.}
\label{tab: abaltion}
\end{table}

\section{Conclusions}
In this paper, we propose a novel approach that automatically animates hair in a still portrait photo 
 without any user  assistance.
An instance-based hair wisp extraction module is proposed to extract hair wisps from an image, which facilitates the animation of hair and helps to generate complex hair motions. To train the instance segmentation model, we construct a hair wisp dataset containing real portrait photos and ground-truth annotations of hair wisps.
Moreover, we introduce a hair wisp animation module that can create realistic motions for hair wisps based on physical models. Benefit from animated hair wisps, our method effectively converts diverse portrait photos containing various hairstyles and  head poses into
high-quality and  high-resolution videos, but also enable the generated videos to  provide an aesthetically-pleasing viewing experience without noticeable artifacts.

\section*{Acknowledgement}
This work was supported by the KAUST Office of Sponsored Research through the Visual Computing Center (VCC) funding,  as well as, the SDAIA-KAUST Center of Excellence in Data Science and Artificial Intelligence (SDAIA-KAUST AI).

{\small
\bibliographystyle{ieee_fullname}
\bibliography{egbib}
}

\end{document}